\pdfoutput=1

\documentclass[11pt]{article}

\usepackage[final]{acl}

\usepackage{times}
\usepackage{latexsym}

\usepackage[T1]{fontenc}

\usepackage[utf8]{inputenc}

\usepackage{microtype}

\usepackage{inconsolata}

\usepackage{booktabs}
\usepackage{graphicx}
\usepackage{adjustbox}
\usepackage{multicol}
\usepackage{multirow}
\usepackage{comment}
\usepackage{subcaption}
\usepackage{xcolor}
\usepackage{amsmath}

\title{What is an ``Abstract Reasoner''? Revisiting Experiments and Arguments about Large Language Models}

\author{Tian Yun \\
  Brown University \\
  \texttt{tian\_yun@brown.edu} \\\And
  Chen Sun \\
  Brown University \\
  \texttt{chensun@brown.edu} \\\And
  Ellie Pavlick \\
  Brown University \\
  \texttt{ellie\_pavlick@brown.edu} \\  
}

\begin{document}
\maketitle

\begin{abstract}
    Recent work has argued that large language models (LLMs) are not ``abstract reasoners'', citing their poor zero-shot performance on a variety of challenging tasks as evidence. We revisit these experiments in order to add nuance to the claim. First, we show that while LLMs indeed perform poorly in a zero-shot setting, even tuning a small subset of parameters for input encoding can enable near-perfect performance. However, we also show that this finetuning does not necessarily transfer across datasets. 
    We take this collection of empirical results as an invitation to (re-)open the discussion of what it means to be an ``abstract reasoner'', and why it matters whether LLMs fit the bill.\footnote{Code and resources are available at: \url{https://github.com/tttyuntian/abstract_reasoner_llm}} 
\end{abstract}

\section{Introduction}

The question of whether large language models (LLMs) are ``abstract reasoners'' has been the frequent subject of recent work, both directly \citep{hu2023context, webb2023emergent, gendron2024large, musker2024semantic} and indirectly \citep{chollet2019measure, mitchell2023comparing, moskvichev2023conceptarc}. The answer to this question feels weighty. LLMs currently dominate modern approaches to AI, and abstract reasoning is arguably the linchpin of general and flexible intelligence \citep{gentner2001introduction, han2024inductive, mitchell2021abstraction}. If LLMs are not abstract reasoners, it follows that fundamental changes are needed in how AI is developed.

The challenge with this question is that there is little consensus on what it means to be an ``abstract reasoner'', and what evidence would convincingly demonstrate that an LLM, or any model, is or is not one.
Most recently, it has been argued that LLMs are not abstract reasoners on the basis of their poor performance when tested out-of-the-box on adapted visual, analogical, and quantitative reasoning tasks (Figure \ref{fig:abstract_reasoning_examples} for examples) that require models to infer and generalize patterns from a limited number of observations \cite{gendron2024large, mitchell2023comparing, stevenson2024can}. 
In this work, we revisit this experimental setup. We replicate the results of earlier studies, but add additional experiments which demonstrate the need for  more nuance before drawing strong conclusions. 

Specifically, we follow the experimental design from \citet{gendron2024large}, and replicate their finding that off-the-shelf pretrained LLMs perform badly across a range of challenging reasoning tasks (\S\ref{sec:frozen_pretrained_llm}). 
However, we find that optimizing just the embedding layer for the task (leaving the transformer blocks frozen) all but eliminates the problem, allowing the model to perform comparably to finetuning the entire model, and sometimes even solve the task perfectly (\S\ref{sec:embedding_finetuning}).
This result extends beyond simple embeddings and, in fact, a frozen pre-trained LLM can perform well on visual reasoning tasks as long as the visual encoder is fine-tuned on in-domain task data (\S\ref{sec:visual_encoder_finetuning}). 

Together, these results paint a more subtle picture of LLMs: much of their representations and inferential capabilities appear to be transferable across very diverse tasks, but non-trivial effort is required on the input side for each new task in order to harness these capabilities. In light of this, we (re-)open the larger discussion which is simultaneously empirical and philosophical (\S\ref{sec:discussion}): What does it mean to be an abstract reasoner, and why do we care whether LLMs fit the bill?

\begin{figure*}
    \centering
    \includegraphics[width=1\linewidth]{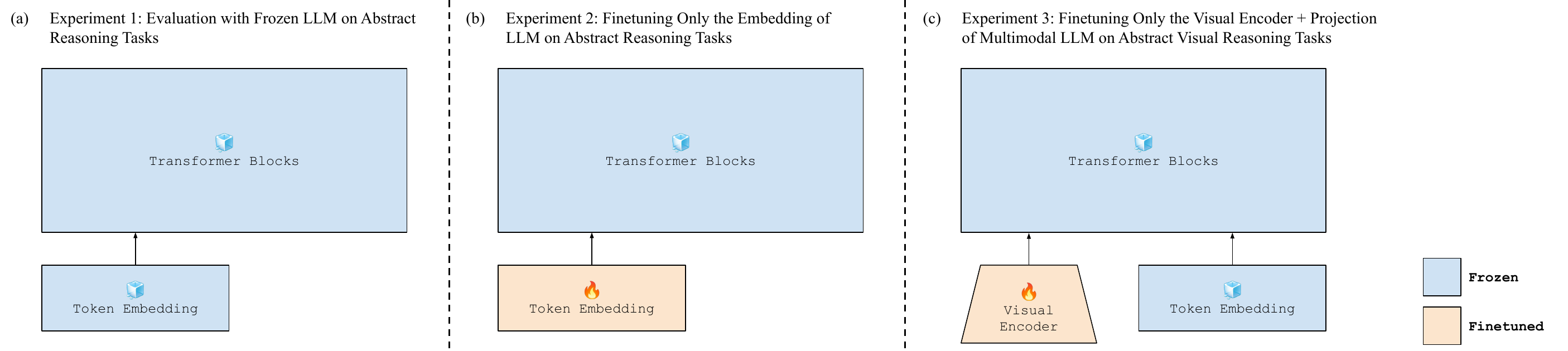}
    \caption{Illustration of our experimental settings. In Setting (a), we freeze the whole LLM and run evaluations. This is treated as language baseline when image captions are inputs on abstract visual reasoning tasks. In Settings (b) and (c), we freeze the pretrained transformer blocks and finetune only the input layers (i.e., token embedding layer and visual encoder). In Setting (c), we freeze the token embedding layer to study the impact of tuning the visual encoder in a controlled setting. Note that the inputs are pure language in Settings (a) and (b), while the inputs are language prompts with image representations in Setting (c).}
    \label{fig:illustration}
\end{figure*}

\section{Related Work}
\label{sec:related_works}

\subsection{Analogical Reasoning}

Prior work has studied the question of abstract reasoning of LLMs via analogical reasoning, such as matrix reasoning \cite{webb2023emergent}, letter-string analogies \cite{mitchell2021abstraction, hofstadter1995copycat} and pointer-value retrieval \cite{zhang2021pointer}.  These analogical reasoning benchmarks require a model to infer the patterns from a limited number of observations and apply the discovered patterns to the new queries.

Despite the impressive performance of LLMs, there is yet no consensus on whether LLMs are strong analogical reasoners. Some studies show evidence suggesting that LLMs can even surpass the human baseline on analogical reasoning tasks \cite{hu2023context, webb2023emergent}, while the others show that LLMs achieve very limited performance on a set of analogical reasoning benchmarks \cite{gendron2024large} or they are not robust to counterfactual examples or irrelevant information \cite{lewis2024using, musker2024semantic}. 
We use similar tasks and models as the prior work, but incorporate additional tasks and a wider range of finetuning experiments in order to situate the results within a larger discussion about abstract reasoning.

\subsection{Visual Analogical Reasoning}
Analogical reasoning can go beyond symbols and words and involve visual input, such as in ARC \cite{chollet2019measure}, ACRE \cite{zhang2021acre}, RAVEN \cite{zhang2019raven, hofstadter1995copycat} and MEWL \cite{jiang2023mewl}. 
Recent approaches on visual analogical reasoning can be categorized into neuro-symbolic methods~\cite{mao2019neuro,hudson2019learning}, or neural networks with implicit representations~\cite{ding2021attention,sun2024does,bhattacharyya2023look}. 
Both approaches roughly follow the same outline of the perception stage and the reasoning stage. 
The perception stage usually relies on task-specific visual encoders, such as symbolic object encoders \cite{zhang2021acre}, object detectors \cite{ding2021attention}, or on task-specific training strategies for these visual encoders \cite{sun2024does, bhattacharyya2023look}. 
The reasoning stage introduces inductive biases by developing task-specific reasoning modules \cite{hu2021stratified, benny2021scale}. In this work, we investigate if the transformer blocks of a pretrained LLM can be used as a reasoner for different visual analogical reasoning tasks.

\subsection{Multimodal Large Language Models}
Prior work shows that transformer blocks pretrained on natural language can be transferred to non-language sequence modeling problems by optimizing new input and output layers \citep{lu2022frozen}. With the rise of LLMs, recent work freezes pretrained vision models and pretrained LLMs, and only learns a mapping to project visual representations to language latent space in order to perform on multimodal tasks \citep{merullolinearly, liu2023visual, li2023blip, liu2024improved}. \citet{tong2025cambrian} investigates the impact of vision-only models in multimodal LLMs and reaches impressive performance on downstream tasks. Our work is similar to these models, but connects it to a larger, more philosophical debate about the meaning of ``abstract reasoning''.

\begin{figure*}[ht]
    \centering
    \includegraphics[width=1.0\linewidth]{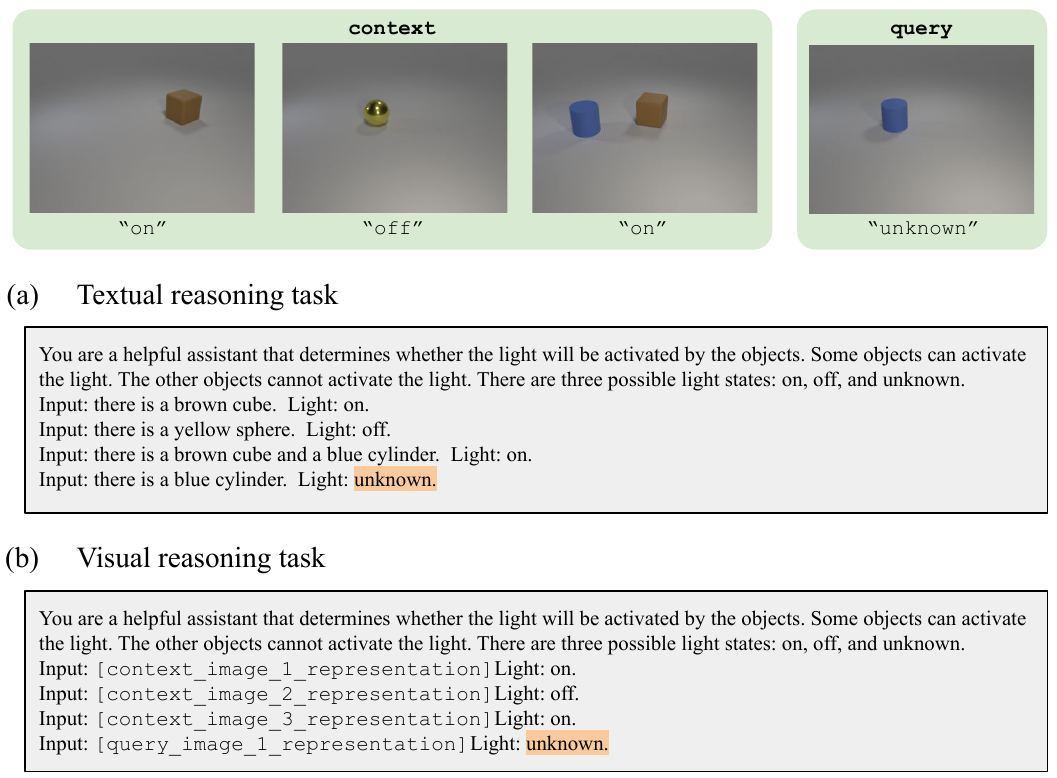}
    \caption{Illustration of the use of language models for text-based and image-based versions of ACRE. Each data example will be formulated into a prompt for an LLM to make a \colorbox{orange!40}{prediction} for the query. In textual reasoning task, each context frame is represented by a frame caption. In visual reasoning tasks, each context frame is represented by an encoded frame representation.} \label{fig:illustration_of_llm_for_reasoning_tasks}
\end{figure*}

\section{Datasets} 
\label{sec:datasets}
\subsection{Reasoning Tasks from Gendron et al.}
We follow the evaluation benchmark used by \citet{gendron2024large} to quantitatively measure the so-called ``abstract reasoning'' capabilities of language models. This benchmark contains seven tasks, each of which evaluates the ability of a model to infer patterns from a limited number of examples. These seven tasks can be divided into two categories: open question answering (OPQA) and multiple-choice question answering (MCQA). OPQA tasks require a model to generate the correct answer, while MPQA tasks require a model to select the correct answer from the given set of answer candidates. OPQA tasks include Abstract Reasoning Challenge (ARC) \cite{chollet2019measure}, BIG-Bench dataset (BBF) \cite{rule2020child, srivastava2022beyond}, Evals-P \cite{achiam2023gpt}, and Pointer-Value Retrieval (PVR) \cite{zhang2021pointer}. MCQA tasks include ACRE$^T$ \cite{zhang2021acre}, RAVEN$^T$ \footnote{ACRE$^T$ and RAVEN$^T$ are text-based version of the original tasks.}\cite{zhang2019raven}, and Evals-P \cite{achiam2023gpt}. For ACRE$^T$ and RAVEN$^T$, we also consider ACRE$^T$-\text{Symb} and RAVEN$^T$-\text{Symb}, where the panel descriptions are converted into symbols (e.g., using integers to represent different objects).

\subsection{Additional (Visual) Reasoning Tasks}
In addition to the models and tasks considered by \citet{gendron2024large}, we additionally consider how well LLM representations transfer fo the multimodal language model framework (MLLM). To support these experiments, we consider two visual reasoning tasks: ACRE \cite{zhang2021acre} and MEWL \cite{jiang2023mewl}. In ACRE, given the 5 context frames and 1 query frame, a model needs to predict the activation status of Blicket detector in the query frame, which can be \textit{on}, \textit{off}, or \textit{unknown}. In MEWL, given 6 context frames and 1 query frame, a model needs to understand the meaning of the novel words and select the correct utterance out of 5 options for the query frame.

\begin{table*}
\centering
\begin{adjustbox}{max width=1.0\textwidth}
\begin{tabular}{l|cccccc|ccccc}
\toprule
 & \multicolumn{6}{c|}{OPQA} & \multicolumn{5}{c}{MCQA} \\
 \cmidrule{2-12}
 & ARC & BBF & Evals-S & PVR & RAVEN$^T$ & RAVEN$^T$-Symb & ACRE$^T$-Text & ACRE$^T$-Symb & Evals-P & RAVEN$^T$ & RAVEN$^T$-Symb \\
\midrule
Random & - & - & - & - & - & - & 33.3 & 33.3 & 50.0 & 12.5 & 12.5 \\
LLaMA2-7b-chat (NZ) & 0.5 & 10.8 & 0.0 & 0.0 & 0.0 & 0.1 & 1.4 & 0.3 & 50.0 & 2.6 & 14.9 \\
LLaMA2-7b-chat (Ours) & 1.0 & 26.4 & 0.0 & 21.8 & 0.0 & 1.0 & 26.4 & 38.1 & 52.0 & 12.9 & 11.4 \\
\bottomrule
\end{tabular}
\end{adjustbox}
\caption{Performance of frozen pretrained LLMs on open question answering (OPQA) and multiple-choice question answering (MCQA) benchmarks. We show the results \texttt{LLaMA2-7b-chat(NZ)} reported in \citet{gendron2024large} and our reproduced results \texttt{(Ours)} following the evaluation from \citet{gendron2024large}.}
\label{tab:pretrained_llm_opqa_mcqa}
\end{table*}

\begin{figure*}
    \centering
    \includegraphics[width=1\linewidth]{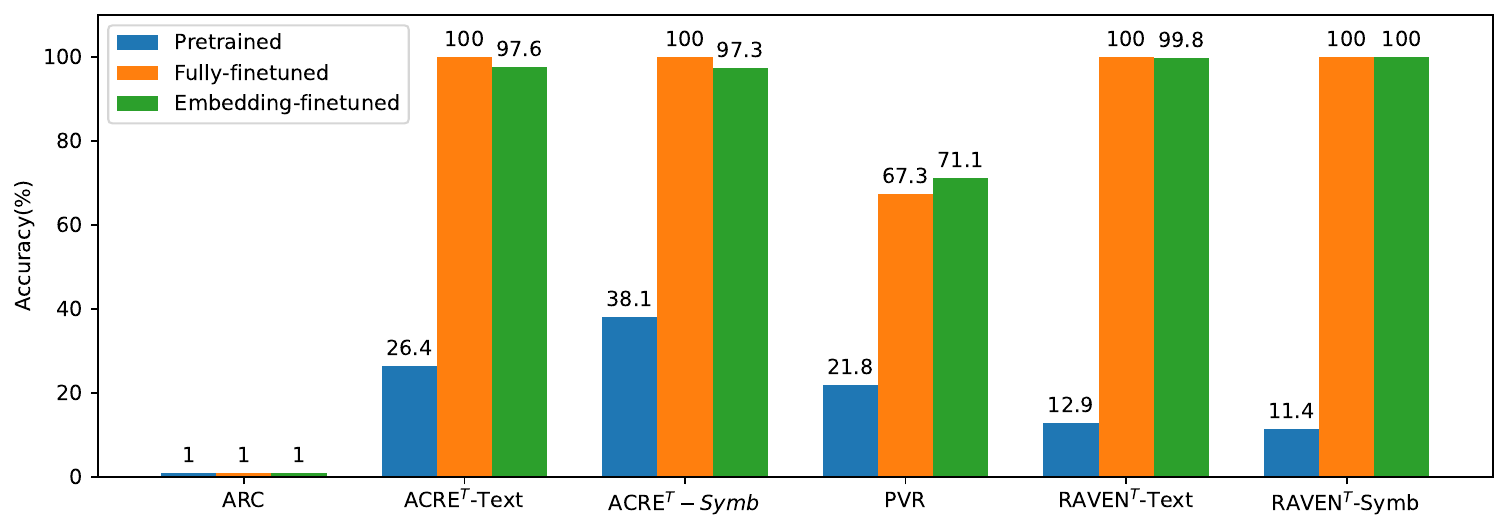}
    \caption{Performance of finetuned LLMs on OPQA (ARC and PVR) and MCQA (ACRE$^T$ and RAVEN$^T$) benchmarks. LLMs with finetuned embedding layer perform significantly better than their pretrained counterparts, and perform on par with or even surpass the fully finetuned LLMs with LORA. Note that ACRE$^T$ and RAVEN$^T$ are text-based version of original datasets, which may make the tasks easier to solve.}
    \label{fig:finetuned_llm}
\end{figure*}

\section{Frozen Pretrained LLMs}

\label{sec:frozen_pretrained_llm}
We first seek to replicate \citet{gendron2024large}'s finding that frozen pretrained LLMs achieve low performance across a large suite of reasoning tasks. We reproduce these evaluations on LLaMA2 with 7 billion parameters \cite{touvron2023llama}. Table \ref{tab:pretrained_llm_opqa_mcqa} shows the results on OPQA and MCQA tasks. We observe that even though there are small gaps between the original results and the reproduced results, the performance of the pretrained LLMs are still low.  
Even when the answer candidates are provided in MCQA tasks, the models mostly perform as poor as random baselines (e.g., 33.3\% on ACRE and 12.5\% on RAVEN). 
We observe significant gaps between original results and ours on BBF and PVR, and attribute them to the choice of parser used to process the model's predictions.

Overall, our results are, if anything, stronger than what has been previously reported in this evaluation setting. But even so, it is hard to argue that these numbers represent ``strong'' performance. We thus agree with \citet{gendron2024large} that these results indicate poor transfer ability. What requires additional investigation, however, is whether this poor transfer is interpretable as a lack of abstract reasoning ability.

\section{Finetuned Embedding Layers} 
\label{sec:embedding_finetuning}
Given that pretrained LLMs perform poorly off-the-shelf, it is natural to ask whether they can be adapted to these task, and if so, just how much adaptation is necessary. 
We explore two ways to finetune the LLMs: (1) finetuning all layers with low-rank adaptation (LoRA) \cite{hu2021lora}; (2) finetunning only the embedding layer of the LLMs. 
LoRA finetuning has become a standard way of adapting a model to a task and represents an upper bound on how well the model could be made to perform the task under the most permissive conditions. 
In contrast, finetuning just the embedding layer represents a conceptually different type of transfer with respect to the question of this paper. Namely, finetuning just the embeddings is analogous to changing just the input to the system--e.g., ensuring the input is in the format the system expects--but leaving the system itself unchanged (see additional discussion and qualifications about this analogy in \S\ref{sec:discussion}). 

We finetune the embedding layer for 50 epochs using AdamW optimizer \citep{loshchilovdecoupled} with early stopping based on the validation set. Following \citet{gendron2024large}, we conduct experiments on 2 OPQA tasks (ARC, PVR) and 2 MCQA tasks (ACRE$^T$, RAVEN$^T$-mcqa\footnote{We are aware of the defects of RAVEN, and we use the original RAVEN since it was previously used by \citet{gendron2024large}}.). 

\begin{figure*}[hbt!]
    \centering
    \includegraphics[width=1.0\linewidth]{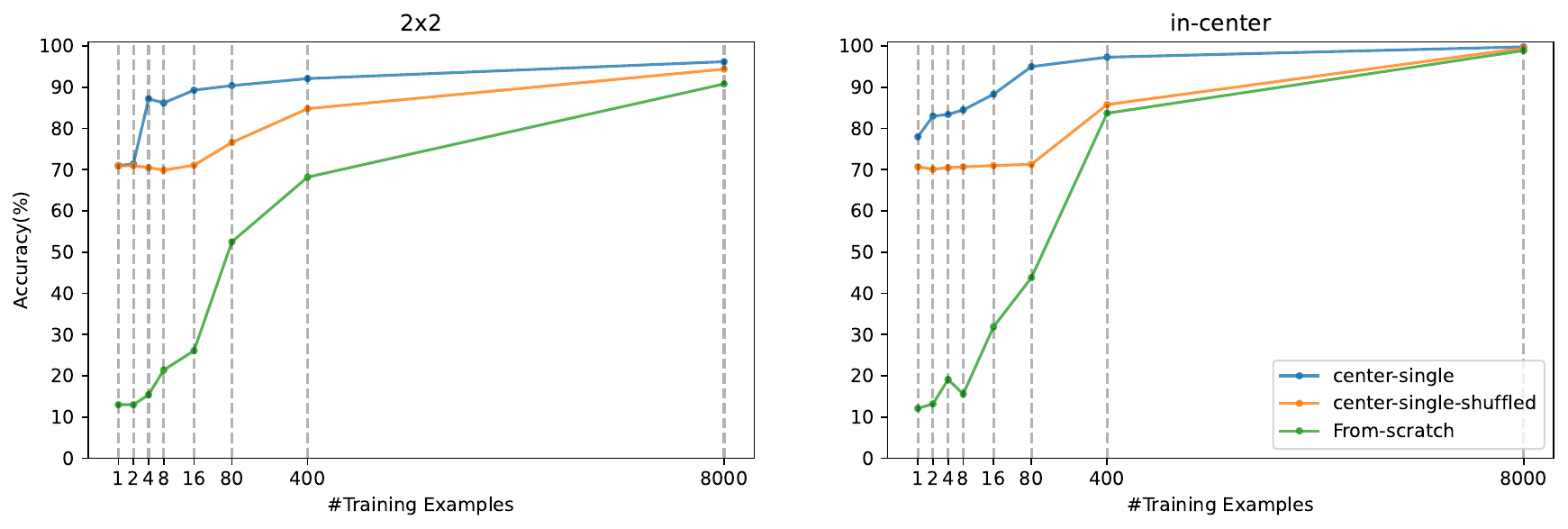}
    \caption{Data efficiency analyses on LLaMA2-7b with token embedding layer finetuned on \texttt{center-single} or \texttt{center-single-shuffled} and further finetuned on \texttt{2x2} and \texttt{in-center} tasks in RAVEN with limited amount of data.
    Y-axis (\texttt{\#Training Examples}) represents the absolute number of examples used for finetuning.
    \texttt{From-scratch} means the token embedding of a pretrained LLaMA2-7b is directly finetuned on \texttt{2x2} and \texttt{in-center} tasks.
    Given that there are 8000 training examples in total, we observe that after finetuning on \texttt{center-single}, the model becomes significantly more data efficient. 
    By comparing \texttt{center-single} and \texttt{center-single-shuffled} lines, we observe that data efficiency of the model mainly comes from the occurrences of task-relevant tokens, rather than the reasoning logic of the tasks.
    }
    \label{fig:data_efficiency}
\end{figure*}

Figure \ref{fig:finetuned_llm} shows the results of finetuned LLaMA2. We observe that LoRA-finetuned models perform significantly better than their pretrained counterparts, and can even solve ACRE$^T$ and RAVEN$^T$ perfectly. Moreover, LLaMA2 with a finetuned embedding layer can perform on par with the LoRA-finetuned LLaMA2\footnote{We attribute the low performance of ARC to its complexity and the length of each data sequence (excluding the expected answer), where 75\% of data has >2000 tokens.}.

\paragraph{Generalizability and Data Efficiency.} 
We conduct experiments on RAVEN$^T$ to further look into two properties of the finetuned input layers: generalizability and data efficiency. An ideal abstract reasoner is expected to generalize to novel tasks with limited amount of observations.
We take LLaMA2-7b's token embedding layer finetuned on one task variant (namely, the \texttt{center-single} task)  and further finetune this layer with varying amount of training examples for 500 steps on two different task variants (\texttt{2x2} and \texttt{in-center}), both of which require reasoning over more attributes (e.g., different object alignments). Figure \ref{fig:raven_text_illustration} shows examples of these tasks. 
We consider three settings: (1) ``\texttt{center-single}'', where the token embedding has been finetuned on the original \texttt{center-single} task; (2) ``\texttt{center-single-shuffled}'', where the token embedding has been finetuned on \texttt{center-single} task with randomly shuffled labels. This setting preserves the visual features, but breaks the logical ``reasoning'' structure of the task, and thus serves as a test of how much of the positive transfer is due to low-level visual cues vs.\ higher-level more abstract features; (3) ``\texttt{from-scratch}'', where the token embedding of a pretrained LLaMA2-7b is directly finetuned on \texttt{2x2} and \texttt{in-center} tasks. We use this to study the impact of finetuning on \texttt{center-single} task.

Figure \ref{fig:data_efficiency} shows the results. LLaMA2-7b with token embeddings finetuned just on 80 examples can perform competitively aganist LLaMA2-7b directly finetuned on full dataset (8k examples) of the tasks. The fairly small gap between the \texttt{center-single} and \texttt{center-single-shuffled} lines suggests that the positive transfer is primarily explained by the lower-level visual features rather than the reasoning logic of the tasks.

\begin{table*}
    \centering
    \begin{subtable}[c]{1\textwidth}
        \centering
        \scalebox{0.68}{
            \begin{tabular}{c|l|ccc}
            \toprule
            & Method & I.I.D. & Compositional & Systematic \\ 
            \midrule
            Language & LLaMA2-7b & 26.4 & 26.1 & 29.9 \\ 
            Baseline & GPT-4 & 66.4 & 66.4 & 64.0 \\
            & GPT-4-Turbo & 69.7 & 69.9 & 67.4 \\
            \midrule
            & NS-OPT & 66.3 & 69.0 & 67.4 \\
            & ALOE & - & 91.8 & 93.9 \\
            Existing & IV-CL & \underline{93.0} & 93.2 & \underline{92.6} \\
            Approaches & LRR & - & \textbf{98.2} & \textbf{99.2} \\
            & LLaVA-NeXT-Mistral-7B & 38.4 & 36.9 & 36.9 \\
            & GPT-4o & 62.6 & 61.5 & 61.7 \\
            \midrule
            Ours & LLaMA2-7b-Object & \textbf{95.5} & \underline{97.5} & 86.5 \\
              
            \bottomrule
            \end{tabular}
        }
        \caption{ACRE}
        \label{tab:finetune_visual_encoder_acre}
    \end{subtable}
    \\
    \vspace{5mm}

    \begin{subtable}{1\textwidth}
        \begin{center}
        \scalebox{0.68}{
            \begin{tabular}{c|l|ccccccccc|c}
                \toprule
                & Method & \texttt{shape} & \texttt{color} & \texttt{material} & \texttt{object} & \texttt{composite} & \texttt{relation} & \texttt{bootstrap} & \texttt{number} & \texttt{pragmatic} & Avg. \\
                \midrule
                Language & LLaMA2-7b & 49.7 & 61.2 & 52.5 & 73.8 & 35.2 & 19.2 & 29.5 & 21.8 & 22.2 & 40.6 \\
                baselines & BERT$^*$ & \underline{94.8} & \underline{98.8} & 97.5 & 19.5 & \textbf{97.8} & \textbf{22.2} & \underline{62.2} & 21.8 & \textbf{99.8} & \underline{68.3} \\
                & GPT-3.5 & \textbf{96.8} & 82.3 & 87.0 & \textbf{98.2} & \underline{88.3} & 20.0 & 45.8 & 22.7 & \underline{26.7} & 63.1 \\
                \midrule
                Existing & ALOE & 34.2 & 33.2 & 31.0 & 19.5 & 30.5 & \underline{21.5} & 27.5 & 23.3 & 20.8 & 26.8 \\
                Approaches & Flamingo-1.1B & 49.3 & 35.3 & 48.5 & 19.2 & 38.2 & 18.8 & 57.3 & \underline{84.2} & 18.0 & 41.0 \\
                \midrule
                Ours & LLaMA2-7b-Object & 59.3 & \textbf{100.0} & \textbf{98.8} & \underline{96.8} & 50.4 & 17.3 & \textbf{87.0} & \textbf{99.5} & 19.2 & \textbf{69.8} \\
                \bottomrule
            \end{tabular}
        }
        \caption{MEWL}
        \label{tab:finetune_visual_encoder_mewl}
        \end{center}
    \end{subtable}
    
    \caption{Results of LLaMA2-7b with train-from-scratch visual encoders on sub-tasks in ACRE and MEWL. \textbf{Bolded results} are the best results, and \underline{underlined ones} are the second best. All language baselines are frozen, except BERT which is finetuned on MEWL tasks. The results show that frozen LLaMA2 with learned visual encoder perform significantly better than its language counterpart and even outperform the existing approaches.}
    \label{tab:abstract_visual_reasoning_main}
\end{table*}

\section{Visual Encoder Trained from Scratch} 
\label{sec:visual_encoder_finetuning}

Prior work has shown that transformer blocks pretrained on natural language can be tranferred to non-language sequence modeling tasks, such as image recognition and protein fold prediction \citep{lu2022frozen}. Given the surprising effectiveness of finetuning just the embedding layer of LLaMA2 on text-only abstract reasoning tasks, we hypothesize that the frozen transformer blocks of a pretrained LLM will perform well on abstract visual reasoning tasks if the visual encoder is tuned for the task. 
That is, we follow the multimodal LLM framework (MLLM) which consists of a visual backbone, a language backbone, and a linear projection layer which maps visual representations to language latent space. We keep the transformer blocks and the token embedding layer of language backbone frozen, and only train the visual encoder and the projection layer. 
If this MLLM with a trained visual encoder can perform better than its language backbone with oracle visual perception, then it provides further evidence for the above interpretation of the frozen LLM as a highly transferable system.

\subsection{Variants of Image Inputs}

In order to run these experiments, we consider three variants of image inputs. Figure \ref{fig:variants_of_image_input} shows the examples of each variant.

\paragraph{Symbol.} A frame is represented by a set of multi-hot object representations, where each object representation is the concatenation of its one-hot vectors for object attributes (i.e., color, material, and shape) and a vector of object location information. This mimics the experiments in \S\ref{sec:embedding_finetuning} by assuming oracle visual perception, and allows us to directly contrast language and visual inputs. 

\paragraph{Object.} A frame is represented by object representations, where each object is an object crop from the frame. This variant assumes ground truth object detection exists in order to control the factors of reasoning performance.

\paragraph{Image.} A frame is represented by its RGB image. This variant simplifies the inputs the most, but requires the visual encoder to encode object properties and spatial relationships between objects directly from the frames. 

\subsection{Language Baseline}
For our language baseline, we provide a frozen LLM directly with language descriptions of the abstract visual reasoning problem. Frame captions can be considered as oracle visual perception, where each frame is represented by its caption (e.g., ``There is a blue cylinder and a brown cube.''). 

\begin{table*}
    \centering
    \scalebox{0.58}{
        \begin{tabular}{l|cccc|cccccccccc}
            \toprule
            & \multicolumn{4}{c|}{ACRE} & \multicolumn{10}{c}{MEWL} \\
            \cmidrule{2-15}
            & I.I.D. & Comp. & Sys. & Avg. & \texttt{shape} & \texttt{color} & \texttt{material} & \texttt{object} & \texttt{composite} & \texttt{relation} & \texttt{bootstrap} & \texttt{number} & \texttt{pragmatic} & Avg. \\
            \midrule
            LLaMA2-7b-Image & 75.8 & 77.7 & 71.7 & 75.1 & 35.0 & 99.8 & 57.7 & 26.2 & 32.7 & 19.8 & 31.8 & 45.2 & 21.3 & 41.1 \\
            LLaMA2-7b-Object & 95.5 & 97.5 & 86.5 & 93.2 & 59.3 & 100.0 & 98.8 & 96.8 & 50.4 & 17.3 & 87.0 & 99.5 & 19.2 & 69.8 \\        
            LLaMA2-7b-Symbol (Linear) & 91.0 & 94.9 & 86.8 & 90.9 & 100.0 & 99.8 & 100.0 & 98.0 & 42.5 & 18.0 & 35.0 & 78.2 & 18.3 & 65.5 \\ 
            LLaMA2-7b-Symbol (MLP) & 98.3 & 99.5 & 84.6 & 94.1 & 100.0 & 100.0 & 100.0 & 98.8 & 71.3 & 16.2 & 91.3 & 99.7 & 22.3 & 77.7 \\
            \bottomrule
        \end{tabular}
    }
    \caption{Analysis on the presence of object-centric information. -Symbol rows can be considered as upper bound, since the inputs are symbolic representations of images. The performance gap between -Image and -Object reflects the importance of object-centric inductive bias in abstract visual reasoning tasks.}
    \label{tab:object_centric_experiment}
\end{table*}

\subsection{Implementation Details}
On ACRE, we use the training set with 6K samples, where each sample contains 6 context frames and 4 query frames. Thus, the training set has 24K sequences. On MEWL, we use the training sets of the 9 sub-tasks, each of which involves 600 samples. Thus, the training set has 5400 sequences. 

For the language backbone, we use LLaMA2 with 7 billion parameters \cite{touvron2023llama}. For the visual backbone, to encode image inputs, we use a 2-layer ViT \cite{dosovitskiy2020image} with 4 attention heads and 768-hidden dimensional space; to encode symbolic representations of images, we use a symbolic encoder which encodes object attributes with embedding layers and encodes objects' location information\footnote{Each object location is represented as $[x_1, y_1, x_2, y_2, w, h, w \times h]$} with a linear layer. 

During finetuning, we freeze the language backbone and finetune the visual encoder and the linear projection. We use the AdamW optimizer with a learning rate of $3 \times 10^{-5}$. We finetune the visual backbone for 20 epochs on ACRE, and 40 epochs on MEWL. The batch size is set to 64. 

\subsection{Results}
Table \ref{tab:abstract_visual_reasoning_main} shows the results of LLaMA2-7b with learned visual encoders on ACRE and MEWL. On ACRE, we observe that LLaMA2 with train-from-scratch visual encoders can perform significantly better than their language-only counterpart. These models can even outperform majority of the multimodal state-of-the-art, including IV-CL \cite{sun2024does} and LRR \cite{bhattacharyya2023look}, which are pretrained with video data. On MEWL, we observe the same pattern that LLaMA2 with learned visual encoders can outperform prior state-of-the-art and also the language baselines which assume perfect visual perception. 

In Table \ref{tab:object_centric_experiment}, we further investigate different ways to represent an image. The large performance gap between LLaMA2-7b-Image and -Object (e.g., average of 41.1\% versus 69.8\% on MEWL), indicating that object-centric information is important for the pretrained transformer blocks to better solve abstract visual reasoning tasks.
In all, these results demonstrate that with a frozen language backbone, learning just the visual encoder from scratch can already improve the model's performance on abstract visual reasoning tasks significantly. However, task-specific design choices, such as object-centric representations, would be needed.

\section{Discussion}
\label{sec:discussion}

The question of whether LLMs are ``abstract reasoners'' has consequences for how we understand and thus how we develop increasingly advanced artificial intelligence. The challenge is that there is no consensus for what it means to be an ``abstract reasoner''. In their recent work, \citet{gendron2024large} operationalize abstract reasoning as the ability to transfer zero-shot to a range of complex reasoning tasks. They find that LLMs perform poorly on this evaluation, and thus conclude that they are not abstract reasoners.

In this work, we reproduce \citet{gendron2024large}'s findings, but push back against their interpretation. In particular, we provide new experiments which show that tuning just the embedding layer is remarkably effective. Indeed, across a variety of textual and multimodal tasks, frozen pretrained LLMs can achieve high levels of performance as long as the input representations are adapted sufficiently for each task \footnote{While we argue that input-level finetuning can enable pretrained models to perform well on a range of tasks, we acknowledge that this does not necessarily imply the models have acquired generalized abstract reasoning in a cognitive sense. Rather, it may reflect the alignment of input representations with the pretrained model’s existing capabilities. A more robust theoretical framework would be needed to precisely distinguish between mere representational alignment and true abstraction across domains and tasks.}.

It seems too stringent a criteria to require that that abstract reasoners perform arbitrary tasks on arbitrary inputs without adaptation. By way of counterargument, consider the good old fashioned AI (GOFAI) systems of the 1990s, which typically included symbolic systems internally, e.g., databases implemented in SQL or rules for logical inference implemented in PROLOG. By most intuitive definitions, these databases and rules would be considered ``abstract'' and the tasks the systems performed over them would be ``reasoning''. But we would not expect these systems to operate well over a database implemented in MongoDB, or to apply rules defined by Python. Rather, the need to operate on representations of a particular format is a consequence of, not an exception to, the system's abstraction. 

Of course, we don't claim that the internal processing of an LLM is exactly analogous to that of a GOFAI system. Of course, in an LLM, tuning the input embedding layer might do more than simply ``rerepresent'', but rather might encode some task-specific processing as well. But interpreted loosely, the analogy is useful for highlighting how the question of adaptability and transferability relates to the question of abstraction and reasoning.

Indeed, this relationship has been considered in depth by philosophers of AI, long before LLMs. For example, \citet{dennett1997} appeals to transferability in his attempt to describe the difference between human cognition\footnote{Dennet's essay is not about reasoning, but rather about \textit{intentional} systems, or systems that have true ``beliefs'' about the world and act according to them.} and simpler computational systems:
\begin{quote}
    Consider the lowly thermostat...we might agree to grant it the capacity for about half a dozen different beliefs...it can believe the room is too cold or too hot, that the boiler is on or off...and so forth...suppose we \textit{de-interpret} its beliefs and desires, it can believe the A is too F or G...and so forth....by attaching the thermostatic control mechanism to different input and output devices, it could be made to regulate the amount of water in a tank, or the speed of a train for instance...But as systems become perceptually richer and behaviorally more versatile, it becomes harder and harder to make substitutions in the actual links of the system to the world without changing the organization of the system itself. 
   ...There comes to be a two-way constraint of growing specificity between the device and the environment. Fix the device in any one state and it demands a very specific environment in which to operate properly (you can no longer switch it easily from regulating temperature to regulating speed or anything else); but at the same time, if you do not \textit{fix} the state it is in, but just plunk it down in a changed environment, its sensory attachments will be sensitive and discriminative enough to respond appropriately to the change...
\end{quote}
\noindent Although Dennett is not discussing the notion of ``abstract reasoners'' \textit{per se}, he observes that intelligent systems do not transfer well unless they are allowed to adapt\footnote{While our experiments adapt the input layer (e.g., token embedding) of a model, adaptation does not have to be limited to the input layers. Indeed, adaptation throughout the model would be consistent with Dennett's argument. A full exploration of this is beyond the scope of this paper, but is an interesting direction for future work.}. Indeed, Dennett argues that this is a defining property, one that differentiates human-like intelligence from simpler (albeit perhaps more abstract) systems such as thermostats.

Dennett's argument is relevant here not because LLMs are human-like or even human-level in their reasoning abilities (they are far from it!). Rather, Dennett articulates a position that is implicit in contemporary discussions about LLMs and ``abstract reasoning''. That is, that we care about how well a system adapts to new environments because adapting well to new environments is a hallmark of intelligence. Indeed, this is often cited explicitly as the motivation for studies of this nature (e.g., ``the question of whether or not LLMs can perform human-like reasoning remains open...'' \cite{gendron2024large}). But if evaluating human-likeness or human-levelness is the motivation for studying abstract reasoning, then arguments such as Dennett's provide a compelling case against using zero-shot transfer ability as a relevant metric.

Of course, there is another, more practical, argument for why we might care about whether LLMs are abstract reasoners, which is simply that we want LLMs to transfer well zero-shot to many tasks in order to facilitate easier, cheaper, and more efficient development of systems. Indeed, the thermostat's highly abstract design is a feature, not a bug. This type of hardware abstraction is what allows similar components and control mechanisms to be readily repurposed to support many types of use cases. A ``human like'' thermostat might be very undesirable.

Thus, before seeking to answer the question of whether LLMs are ``abstract reasoners'', we must first determine, as a community, why we care. Do we care because we want to understand how human-like they are, or do we care because we want to facilitate more efficient technological progress? Almost certainly, we care about both, but we should not expect the same experiments to bear on both lines of inquiry. Finding clarity around these questions--what is an abstract reasoner and why do we care about building one?--is the essential next step if we are to make progress toward either, or both, goals. 

\section{Conclusion}
In this paper, we have (re-)opened the discussion of what it means to be an ``abstract reasoner'', and why it matters whether LLMs are ``abstract reasoners''. We have offered empirical results showing that off-the-shelf pretrained LLMs indeed perform poorly on reasoning benchmarks in a zero-shot setting. However, on a variety of textual and multimodal reasoning tasks, frozen pretrained LLMs can reach high levels of performance when the input embeddings are tuned. With this collection of empirical results, we argue that there is a need to determine why we care about whether LLMs are ``abstract reasoners'' before answering this question.

\section{Acknowledgement}
We would like to thank all reviewers and the area chair for their valuable feedback. 
We would like to thank Samuel Musker, Calvin Luo, and other members of the SuperLab at Brown University for their discussions and insights. The project depicted is sponsored in part by a Young Faculty Award from the Defense Advanced Research Projects Agency, Grant \#D24AP00261. The content of the information does not necessarily reflect the position, or the policy of the government and no official endorsement of this work should be inferred.

\bibliography{custom}

\newpage
\appendix
\counterwithin{figure}{section}
\counterwithin{table}{section}

\section{Limitations}
Since the experiments are compute-intensive, our experiments mainly focus on LLaMA2-7b, but there are many other LLMs trained with different number of parameters, data, or inductive biases. We also only consider one prompt template for each reasoning task, and acknowledge that experimenting with more prompts can provide a more comprehensive evaluation of pretrained LLMs. Last, we use parsers to parse the predictions of models in order to compare with the labels. One alternative approach is the use of other LLMs to compare the predictions with the labels. Some of the above concerns are common challenges for existing evaluation of LLMs. Future research could run evaluations on more LLMs and explore whether the tuning other layers (e.g., output layer, middle layers of transformer blocks) can lead to performance improvement, further proving that LLMs need some amount of task adaptations.

\section{Additional Figures}
We show additional figures to illustrate the reasoning tasks we considered and variants of image inputs.

\begin{figure*}
    \centering
    \includegraphics[width=1\linewidth]{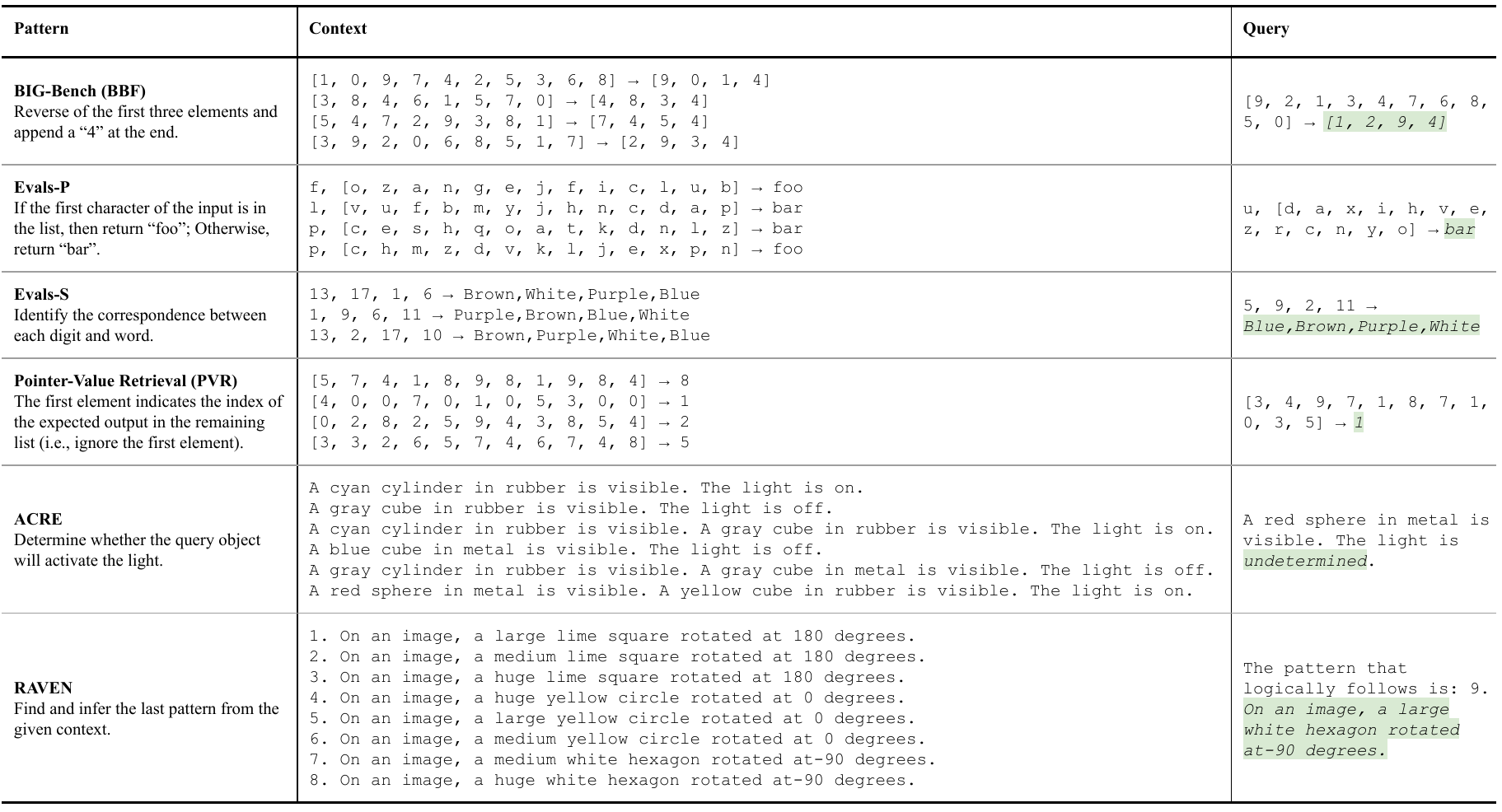}
    \caption{Data examples of abstract reasoning tasks. }
    \label{fig:abstract_reasoning_examples}
\end{figure*}

\begin{figure*}[ht]
    \centering
    \includegraphics[width=.8\linewidth]{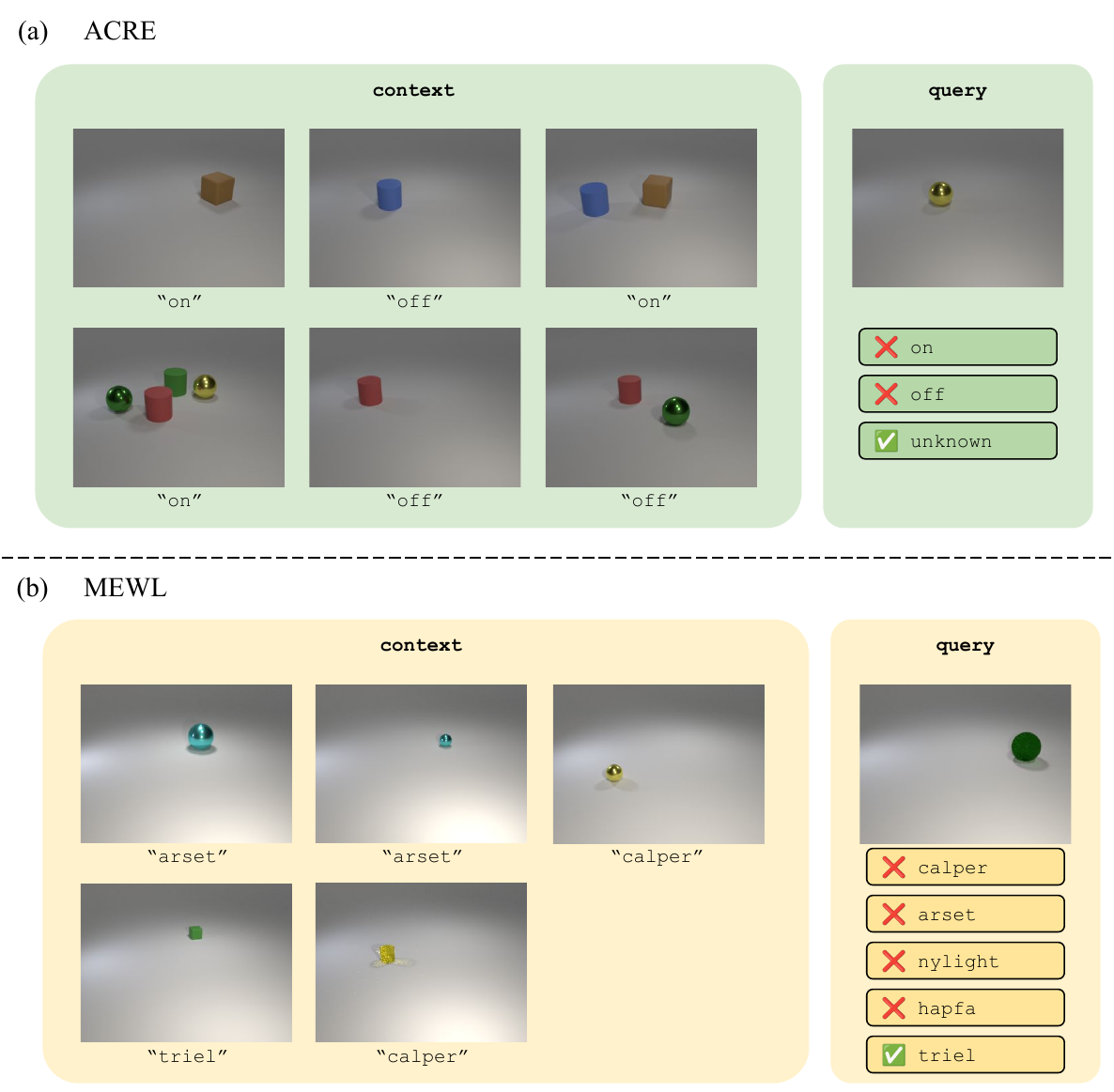}
    \caption{Data examples of abstract visual reasoning tasks.
    }
    \label{fig:abstract_visual_reasoning}
\end{figure*}

\begin{figure*}[hbt!]
    \centering
    \includegraphics[width=0.8\linewidth]{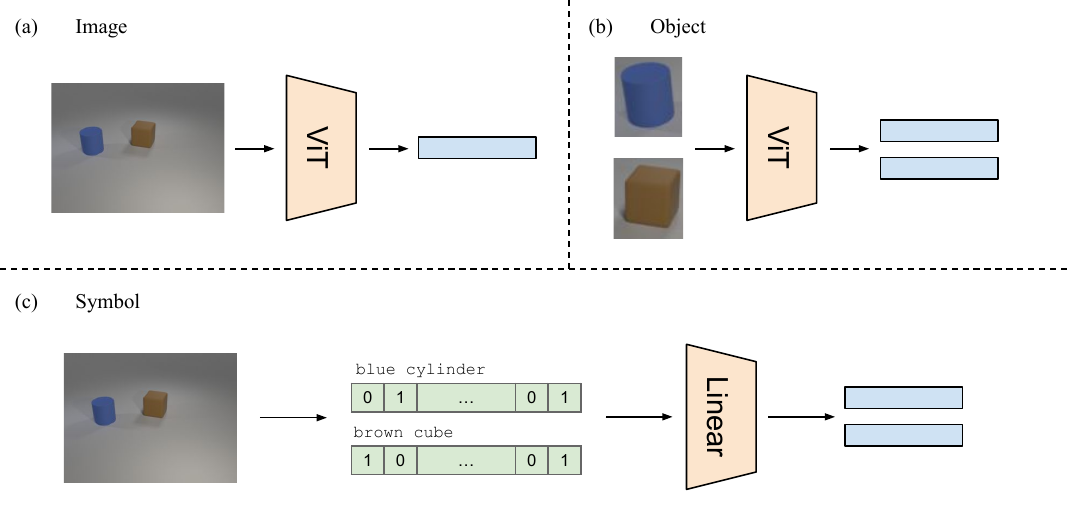}
    \caption{Examples of variants of image inputs. (a) An image is directly fed into a ViT and obtain an image representation. (b) Each object crop is fed into a ViT and obtain an object representation. (c) Each object is parsed into a multi-hot vector, and a linear layer will output a corresponding object representation.}
    \label{fig:variants_of_image_input}
\end{figure*}

\begin{figure*}[hbt!]
    \centering
    \includegraphics[width=0.7\linewidth]{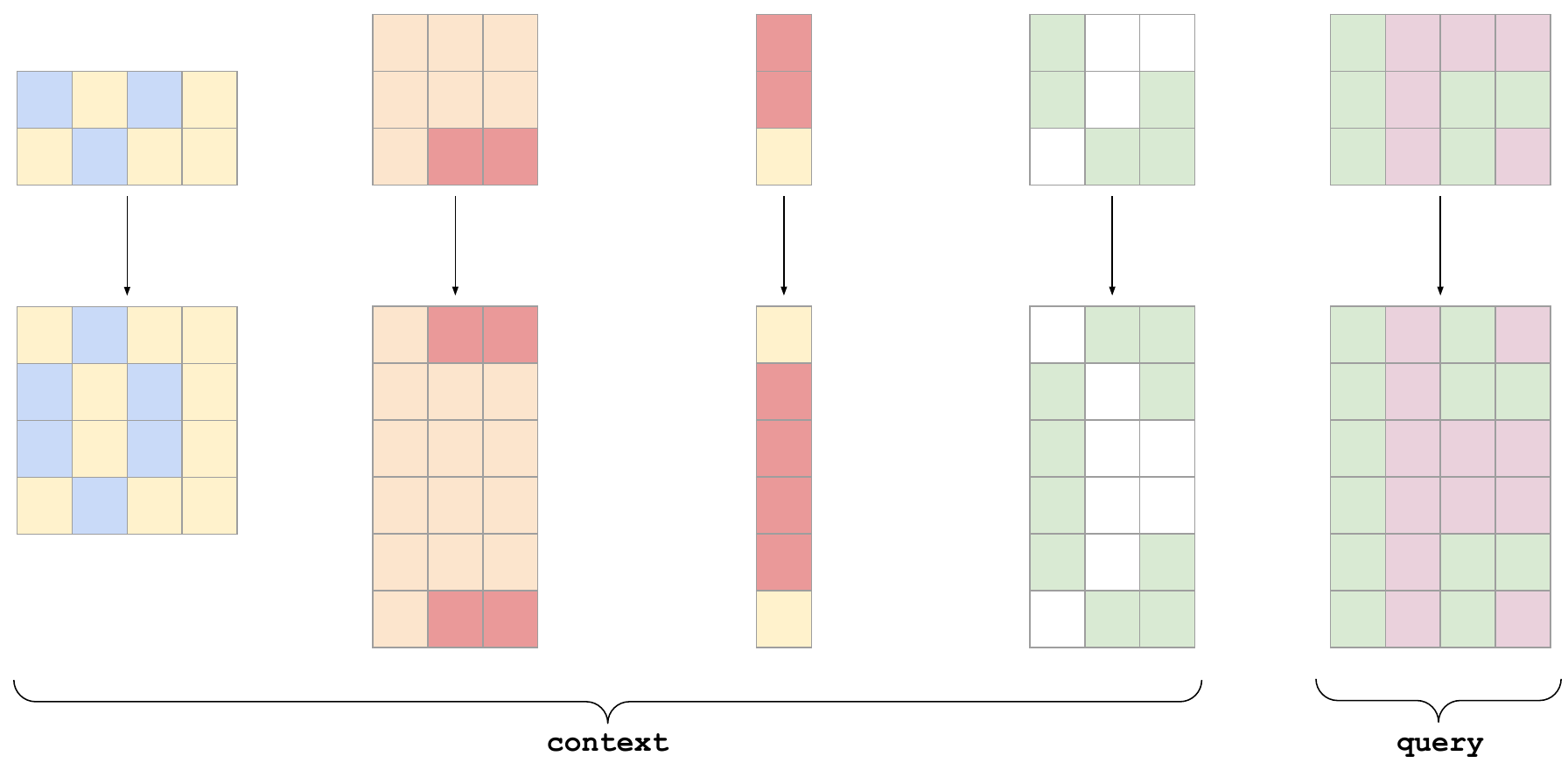}
    \caption{Example of ARC dataset. There are 4 context examples and 1 query, where each example has an input grid (top) and an output grid (bottom). Each grid is represented as an integer array, where each integer refers to a color. In this example, the task is to generate the symmetry of the input grid and stack the symmetry on top of the original input.}
    \label{fig:arc}
\end{figure*}

\begin{figure*}
    \centering
    \includegraphics[width=1.0\linewidth]{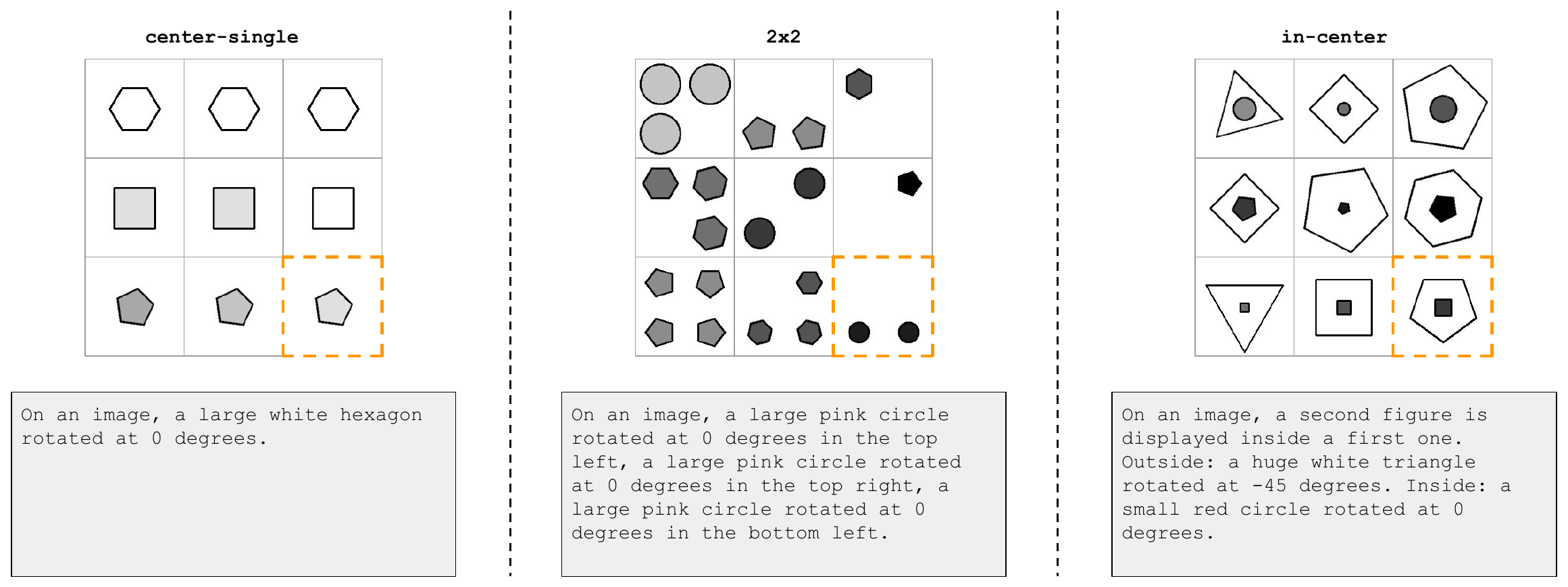}
    \caption{Examples of RAVEN$^T$ tasks used in generalizability and data efficiency analysis. Top shows the data example, and bottom shows the language description of the first frame in each example. The task is to fill in the ninth pattern (highlighted in orange) given the eight context frames. We focus on three tasks: \texttt{center-single}, \texttt{2x2} and \texttt{in-center}. \texttt{center-single} is the simplest task, since there is always only one object in each frame. \texttt{2x2} and \texttt{in-center} consider more than one objects in the frames and also involve different object alignments.}
    \label{fig:raven_text_illustration}
\end{figure*}

\end{document}